\documentclass[twocolumn]{cinc}
\usepackage{graphicx}
\begin{document}
\bibliographystyle{cinc}

% Keep the title short enough to fit on a single line if possible.
% Don't end it with a full stop (period).  Don't use ALL CAPS.
\title{Aortic Pressure Forecasting with Deep Learning}

% Both authors and affiliations go in the \author{ ... } block.
% List initials and surnames of authors, no full stops (periods),
%  titles, or degrees.
% Don't use ALL CAPS, and don't use ``and'' before the name of the
%  last author.
% Leave an empty line between authors and affiliations.
% List affiliations, city, [state or province,] country only
%  (no street addresses or postcodes).
% If there are multiple affiliations, use superscript numerals to associate
%  each author with his or her affiliations, as in the example below.

\author {Eliza Huang$^{1,*}$, Rui Wang$^{1,3,*}$, Uma Chandrasekaran$^{2}$, Rose Yu$^{1,3}$ \\
\ \\ % leave an empty line between authors and affiliation
 $^1$ Northeastern University, Boston MA, USA\\
$^2$  Abiomed, Danvers MA, USA\\
$^3$  UC San Diego, La Jolla CA, USA\\
$^*$  Both authors contributed equally }

\maketitle

% LaTeX inserts the ``Abstract'' heading in the proper style and
% sets the text of the abstract in italics as required.
\begin{abstract}

% Of course, you must insert blank lines
% between the paragraphs of your LaTeX input file, since this is the
% only way to indicate paragraph boundaries.  Make sure that no blank
% lines appear between paragraphs in the formatted output, however.)
% 
% 
    Mean aortic pressure (MAP) is a major determinant of perfusion in all organs systems. The ability to forecast MAP would enhance the ability of physicians to estimate prognosis of the patient and assist in early detection of hemodynamic instability. However, forecasting MAP is challenging because the blood pressure (BP) time series is noisy and can be highly non-stationary.  The aim of this study was to forecast the mean aortic pressure five minutes in advance,using the 25 Hz time series data of previous five minutes as input. 

    We provide a benchmark study of different deep  learning models for BP forecasting. We investigate a left ventricular dwelling transvalvular micro-axial device, the Impella, in patients undergoing high-risk percutaneous intervention.  The Impella provides hemodynamic support, thus aiding in native heart function recovery. It is also equipped with pressure sensors to capture high frequency  MAP measurements at origin, instead of peripherally. Our dataset and the clinical application is novel in the BP forecasting field.  We performed a comprehensive study on time series with increasing, decreasing, and stationary trends. The experiments show that recurrent neural networks with Legendre Memory Unit  achieve the best performance with an overall forecasting error  of 1.8 mmHg.

\end{abstract}
% LaTeX inserts the extra space here automatically.

\section{Introduction}
% Section numbering is automatic.  The examples on the next page
% illustrate how to make subsections.

\begin{figure}[ht]
%\centering
%\includegraphics[width=7.9cm]{impella/Insertion_Impella_CP.png}
\includegraphics[width=9cm]{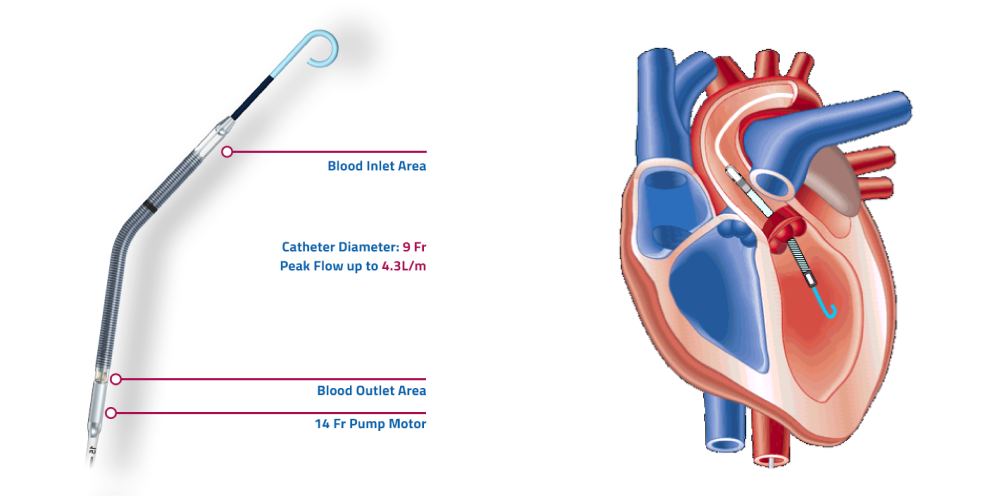}
\caption{Impella CP [Left] and it position [Right].}
\label{Impella}
\end{figure}

Patients with severe multi-vessel coronary artery disease (CAD), unprotected left main coronary artery stenosis, last remaining patent vessel, and severely reduced left ventricular (LV) ejection fraction (EF) are often turned down from cardiac surgery and are increasingly referred for high-risk percutaneous coronary intervention (HRPCI)\cite{russo}. The Impella CP is a left ventricular dwelling transvalvular micro-axial device. As shown in Figure \ref{Impella}, Impella is used during HRPCI to prevent hemodynamic instability and to improve clinical outcomes. The Impella CP entrains blood from the left ventricle and expels into the ascending aorta. The hemodynamic effects of Impella devices include an increase in cardiac output, improvement in coronary blood flow resulting in a decrease in LV end-diastolic pressure, pulmonary capillary wedge pressure, myocardial workload, and oxygen consumption\cite{burkhoff}.

% LaTeX converts the reference tag ('WP-89') into a reference number,
% which it inserts in the formatted text as '[1]'.  If you use BibTeX
% to handle references, BibTeX finds a reference with the tag 'WP-89'
% in 'refs.bib', and inserts that reference into the ``References''
% section in CinC style.
% 
\subsection{Clinical Significance}
Maintenance of a constant mean aortic pressure (MAP) is vital to ensure adequate organ perfusion \cite{chemla}. Studies have shown that  increase in the duration of time spent below MAP threshold of 65 mmHg is associated with worse patient outcomes such as risk of mortality and organ dysfunction \cite{varpula,dunser}. Advance warning of imminent changes in MAP, even if the warning comes only 5 or 15 mins ahead, could aid in prompt management of the patient prior to a total hemodynamic collapse. 
Impella  is equipped with pressure sensors to capture high frequency MAP measures at the origin, instead of peripherally. Impella signals have pressure measurements sampled at a high  frequency that accurately reflect how blood supply is being driven to organs. The  machine learning approach to MAP forecasting highlights the utility of Impella CP not only as a circulatory support but also as a real-time alerting system of cardiac functions, thus obviating the need for additional invasive arterial lines and aiding in prompt escalation of therapy.

 \subsection{Technical Significance}  
 Many researchers have adopted deep learning  for clinical prediction, but mostly for classification tasks \cite{Purushothama, Harutyunyan}. Previously in the PhysioNet Challenge, neural network models have been used on electrocardiogram (ECG) and aortic blood pressure time sequences to classify patients with acute hypotensive episodes in an one-hour forecast window \cite{physionet}. Hatib et al. 2018 \cite{hatib} achieved a sensitivity of 92 percent (AUC 0.97) when classifying arterial hypotension 5 minutes in advance. However, there are limited works  on forecasting aortic pressure with deep  learning models. For example. Peng et al. 2017 \cite{pressure}  took 8-10 minutes of wearable sensor data (ECG, PPG, BP) per day as input and predicted once per day, seeing RMSEs between 1.8-5.81 mmHg for systolic/diastolic BPs over 6 months in a \emph{healthy} patient population.  Our approach combined the medical device type data source like \cite{hatib} with the sequence generation approach in \cite{pressure}.
 
Forecasting MAP is challenging because the aortic pressure time series is highly non-stationary and dynamic. Error and uncertainty are increased for long-term forecasting. Hence, there is a need for a systematic benchmark study of forecasting with deep learning models. 
We aim to apply deep sequence models to forecast the MAP five minutes in advance, using the data from previous five minutes as input. We choose five minutes as the forecasting window to be aware of the changes in continuous aortic pressure within an actionable time frame for clinician intervention.

 \section{Deep Sequence Models} 
% \subsection{Data Description}  
 We benchmark the performances of the following models to predict the MAP time series directly from the input waveforms. By forecasting the MAP values, the model inherently captures ranges of blood pressures- including hypotensive and hypertensive episodes.

% To evaluate all models' performance and their generalization ability, we trained all models on several training sets with different proportions of three types sequences and reported root mean square error (RMSE) on the three test datasets.

\vspace{3mm}
{\textbf{Deep Neural Network (DNN)}} \cite{DNN}: formed by one input layer, multiple hidden layers and one output layer. For our task, we use DNN in an autoregressive manner. We build one DNN with single unit in the output layer to do one step ahead prediction, and keep recursively feeding back the predictions to do multiple steps ahead predictions.

\vspace{3mm}
{\textbf{Recurrent Neural Network (RNN)  w/ Long Short Term Memory (LSTM)}} \cite{RS2S,lstm}:  Sequence to Sequence \cite{RS2S}  maps the input sequence to a fixed-sized vector using an  Encoder, and then map the vector to the target sequence with a Decoder. We use RNNs to retain the sequential information in the time series, as its hidden layer can memorize information processed through the same weights and bias. However, vanilla RNNs have trouble learning long term dependencies. LSTM \cite{lstm} were designed for problems with long term dependencies and also addresses the vanishing gradients issue by using a memory cell state. For the encoder, we use a Bidirectional RNN so that the model can process the data in both the forward and backward directions.  For the decoder, we used another  LSTM to decode the target sequence from the hidden states. 

\vspace{3mm}
{\textbf{RNN with Attention}}\cite{attention}: The attention mechanism learns local information by utilizing intermediate encoder states for the context vectors used by the decoder. It is used to overcome the disadvantage of fixed-length context vector by creating shortcuts between the context vector and the entire source input.

\vspace{3mm}
{\textbf{RNN w/ Legendre Memory Unit (LMU)}}\cite{LMU}: The \texttt{LMU} model \cite{LMU} further addresses the issue of vanishing and exploding gradients commonly associated with training RNNs by using cell structure derived from first principles to project continuous-time signals onto $d$ orthogonal dimensions. By mapping legendre polynomials into a linear memory cell, the LMUs can efficiently handle long-term temporal dependencies and converge rapidly, and use few internal state-variables to learn complex functions. This cell structure enables the gradient to flow across the continuous history of internal feature representations. \texttt{LMU} is a recent innovation that achieves state-of-the-art memory capacity while ensuring energy efficiency, making it especially suitable for and the chaotic time-series prediction task under medical device context.

\vspace{3mm}
{\textbf{Temporal Convolutional Network}}\cite{TCN}: \texttt{Temporal Convolutional Neural Network(TCN)} is a model that has a convolutional hidden layer, operating over a one dimensional sequence. Convolutional neural networks create hierarchical representations over the input sequence in which nearby input elements interact at lower layers while distant elements interact at higher layers. This provides a shorter path to capture long-range dependencies compared to the chain structure modeled by recurrent networks. We used several convolutional blocks followed by a flatten layer and several fully connected layers.

\vspace{3mm}
{\textbf{Transformer}}\cite{transformer}: The \texttt{Transformer} model is the first transduction model relying entirely on self-attention to compute representations of its input and output without using sequence-aligned RNN or convolutions. The encoding component is a stack of encoders and the decoding component is a stack of decoders of the same number. The encoders are all identical in structure and each one is composed of two sub-layers: one multi-head attention layer and one fully connected layer. The decoder has both those layers, but between them is an attention layer that helps the decoder focus on the output of the encoder stack. 

\vspace{3mm}
{\textbf{Convolutional Neural Pyramid}}\cite{pyramid}: Different time series may require feature representations at different time scales. Multiscale models are able to encode the temporal dependencies with different timescales and learn both hierarchical and temporal representations. A cascade of features are learned in two streams: first across different pyramid levels enlarges the receptive field, second learns information in each pyramid level and finally merges it to produce the final result.

 \section{Experiments}
The Impella placement signal tracing provides the aortic pressure (mmHg) signal as measured by an optical sensor located at the outlet of Impella. Motor speed (rotations per minute) provides the Impella pump speed as set on the console. Both signals are captured at 25 Hz. We utilized data from 67 Impella cases. We refer to the 25HZ time series as real time (RT) data. As we are mainly interested in BP trends, we   converted all 25HZ data into 0.1 HZ averaged time (AT) data by averaging every 250 RT data points, which is then used for forecasting.

We used a sliding window approach to generate sequences of length 15,000 samples (10 mins). We categorized these sequences into three types based on trends: increasing (I) sequences, decreasing (D) sequences, and stationary (S) sequences where increasing and decreasing were defined by a swing of 10+ mmHg.  50,705 increasing sequences, 50,577 decreasing sequences and 419,559 stationary sequences were collected.  We trained all models using an RMS-prop optimizer, a learning rate decay of 0.8, and batch size 64. We performed a hyper-parameter search on a 10\% hold-out  data. Additional notes on training details are  in the supplementary material.

 \begin{figure}[ht]
\includegraphics[width=9.0cm]{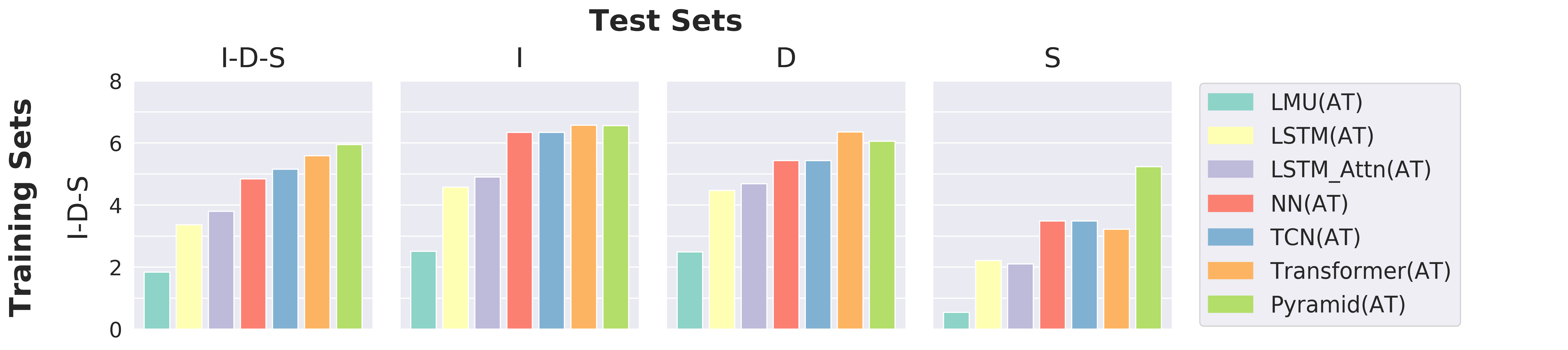}
\caption{RMSE (mmHg) of select models: Each barplot shows models' prediction errors for each  test set: increasing I, decreasing D, stationary S with N=50,000 sequence per set.  "I-D-S" set contains equal proportions of all three types of sequences, N=150,000 sequences.}
\label{distribution}
\end{figure}

We trained all models on datasets with different sequences types (I-D-S). We then tested on dataset for different scenarios: I-D-S, as well as increasing (I), decreasing (D), and stationary (S) individually.  
Figure \ref{distribution} displays the testing errors (RMSEs) comparison for different models. The LMU model achieved the best result, with an average RMSE of 1.837 mmHg on the I-D-S test set. 

 \begin{figure}[ht]
\includegraphics[width=9.0cm]{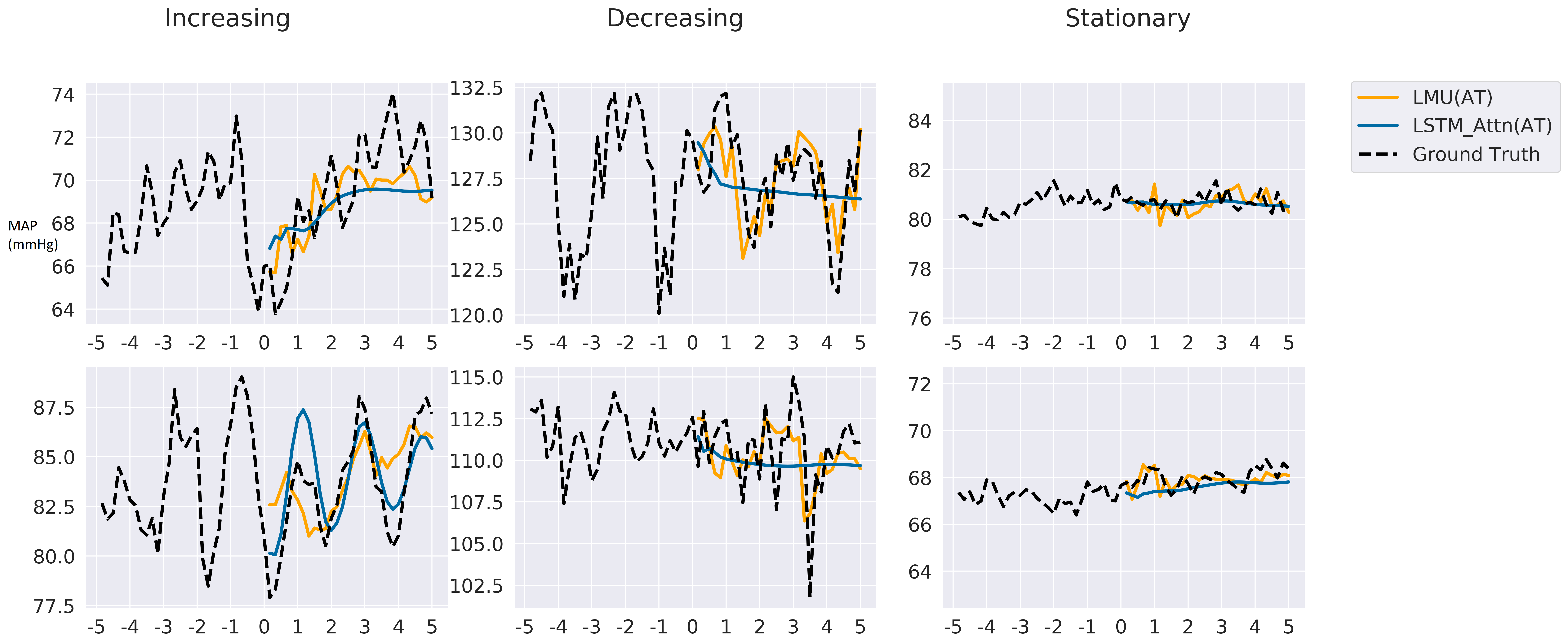}
\caption{Model predictions (from the I-D-S training set).}
\label{preds_aop}
\end{figure}

Figure \ref{preds_aop} show the forecasts generated by top two models, \texttt{LMU} and \texttt{LSTMs with Attention}. The dashed line is the true aortic pressure and the solid lines are the model predictions.  We can see that the forecasting models fit the ground truth quite well. This highlights the LMU's response to MAP fluctuations on a by minute basis.  

\begin{figure}[ht]
\includegraphics[width=9.0cm]{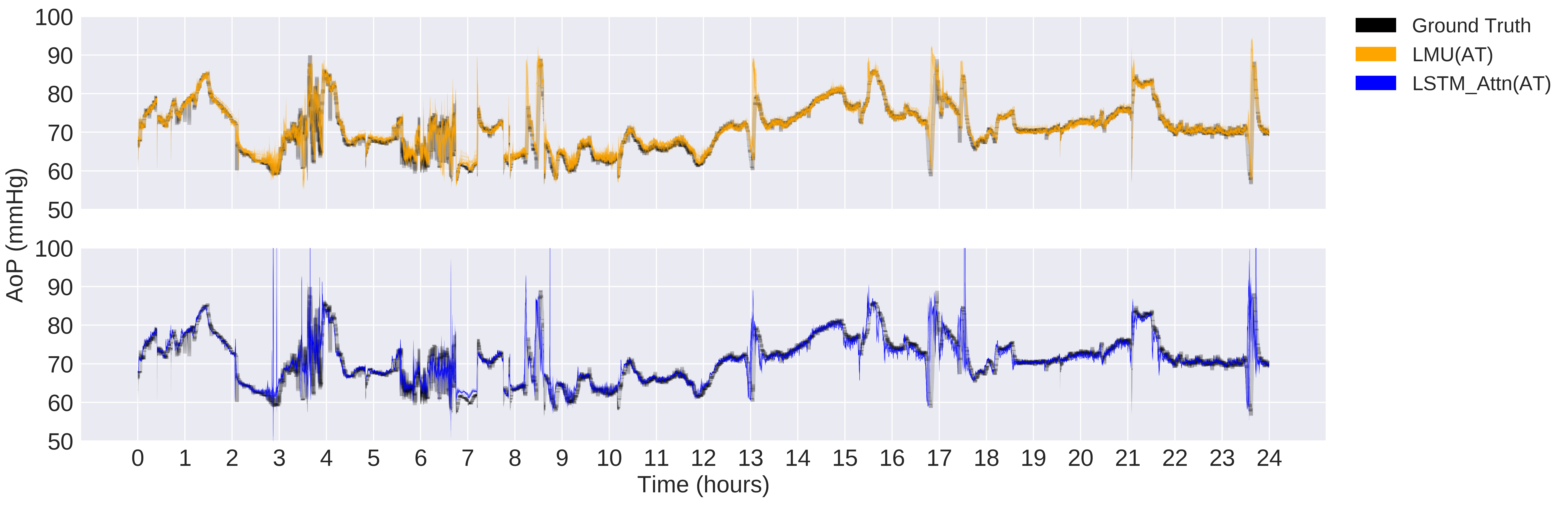}
\caption{Results from top two models (\texttt{LMU} and \texttt{LSTMs with attention}) predictions compared to Ground Truth for one Impella recording over 24hrs}
\label{day_results}
\end{figure}

Figure \ref{day_results} shows the MAP predictions against the ground truth with acute changes for one recording over the course of 24 hours. We can see the acute changes in MAP, both in increasing and decreasing trends, and how well the model follows these trends. We observe that Legendre Memory Units consistently performs the best among all models due its memory capacity and continuous-time model architecture design. 

In the supplementary material, we include the results of all models trained on permutations of the I-D-S sets. For example, we trained the models just using increasing and decreasing samples and tested on I-D-S sets to see if the model could then forecast stationary blood pressure values. We also report all specific RMSE values for training, validation, and test sets for the winning model.

\section{Discussion and Conclusion}   
We explored the possibility of forecasting MAP among patients with Impella support using deep learning. Given the role of MAP as the indicator of perfusion, forecasting the long-term trends in Aortic Pressure would greatly enhance the ability of practitioners to anticipate the condition of their patients on Impella during recovery and during weaning off mechanical circulatory support. 

We prototyped the MAP forecasting model using pump performance data collected by the Impella controller console. This pump performance data captures the pressures the pump experiences and the pump's operating characteristic at aortic pressure origin. We conducted comprehensive experiments with advanced deep sequence  models on the pump performance metrics. The experiments show promising results with the {Legendre Memory Units} (\texttt{LMU}) achieving the best, consistent performance on the task of five minutes ahead mean aortic pressure prediction. 

Our approach demonstrates the promise of deep learning for near real-time blood pressure forecasting.   We investigated many model architectures and achieved an improved  RMSE value of 1.837 mmHg using RNN \texttt{LMU} while  forecasting at a near continuous resolution.  This evaluation determined that deep sequence learning for clinical prediction can accurately forecast  physiologic waveforms with \texttt{LMU} achieving top performance. 

Future work will investigate forecasting out further (10,15 min) and evaluating how error changes as forecast time increases. We will be considering additional patient information, e.g. external intervention (medication) effects and governing equations of physiological features, into the feature set for the RNN Legendre Memory Units model.

\balance

\section*{Acknowledgments}  
This work was sponsored by a research grant from Abiomed, Inc. (Danvers, MA).

% LateX generates the ``References'' heading automatically and switches
% to 9 point type for the bibliography.  Please  use BibTeX and
% follow the examples in the sample 'refs.bib' file to enter your references.
%\bibliography{refs}

\begin{thebibliography}{9}
\bibitem{russo}
Russo G, Burzotta F, D'Amario D, Ribichini F, Piccoli A, Paraggio L, Previ L, Pesarini G, Porto I, Leone AM, Niccoli G, Aurigemma C, Verdirosi D, Trani C, Crea F. \emph{Hemodynamics and its predictors during Impella-protected PCI in high risk patients with reduced ejection fraction}. Int J Cardiol. 274:221-225, 2019. 

\bibitem{burkhoff}
Burkhoff D, Naidu SS. \emph{The science behind percutaneous hemodynamic support: a review and comparison of support strategies}. Catheter Cardiovasc Interv. 80:816-29, 2012.

\bibitem{chemla}
Chemla D, Antony I, Zamani K, Nitenberg A. \emph{Mean aortic pressure is the geometric mean of systolic and diastolic aortic pressure in resting humans}. Journal of Applied Physiology 99:6, 2278-2284, 2005.

\bibitem{varpula}
Varpula M, Tallgren M, Saukkonen K, Voipio-Pulkki L-M, Pettilä V \emph{Hemodynamic variables related to outcome in septic shock}. Intensive Care Med 31:1066–1071, 2005.

\bibitem{dunser}
Dunser MW, Takala J, Ulmer H, Mayr VD, Luckner G, Jochberger S, Daudel F, Lepper P, Hasibeder WR, Jakob SM  \emph{Arterial blood pressure during early sepsis and outcome}. Intensive Care Med 35:1225–1233, 2009.

\bibitem{Purushothama}
Purushothama S, Mengb C, Chea Z, Liu Y. \emph{Benchmarking deep learning models on large healthcare datasets}. Journal of Biomedical Informatics 83, 112-134, 2018.

\bibitem{Harutyunyan}
Harutyunyan H, Khachatrian H, Kale D, Steeg G, Galstyan A. \emph{Multitask learning and benchmarking with clinical time series data}. Scientific Data, doi: 10.1038/s41597-019-0103-9, 2017.

\bibitem{physionet}
Henriques, JH. Rocha, TR. \emph{Prediction of Acute Hypotensive Episodes Using Neural Network Multi-models}. Computers in Cardiology 36:549-552, 2009.

\bibitem{DNN}
Schmidhuber, J. \emph{"Deep Learning in Neural Networks: An Overview".} Neural Networks. 61: 85–117, 2015. 

\bibitem{RS2S}
Ilya Sutskever, Oriol Vinyals, Quoc V. Le. \emph{Sequence to Sequence Learning with Neural Networks.} NeurIPS 2014.

\bibitem{lstm}
Sepp Hochreiter, Jürgen Schmidhuber. \emph{Long Short-Term Memory}. Neural Computation, Volume 9 Issue 8, 1997.

\bibitem{LMU}
Aaron R. Voelker, Ivana Kajic, Chris Eliasmith. \emph{Legendre Memory Units: Continuous-Time Representation in Recurrent Neural Networks.} NeurIPS 2019.

\bibitem{attention}
Minh-Thang Luong, Hieu Pham, Christopher D. Manning \emph{Effective Approaches to Attention-based Neural Machine Translation.} arXiv:1508.04025, 2015.

\bibitem{TCN}
Shaojie Bai, J. Zico Kolter, Vladlen Koltun. \emph{An Empirical Evaluation of Generic Convolutional and Recurrent Networks for Sequence Modeling.} arXiv:1803.01271v2, 2018.

\bibitem{transformer}
Ashish Vaswani, Noam Shazeer, Niki Parmar, Jakob Uszkoreit, Llion Jones, Aidan N. Gomez, ukasz Kaiser, Illia Polosukhin. Sjauw, Vivi Rottschafer, Michel Vellekoop, Paul Zegeling. \emph{Attention Is All You Need.} arXiv:1706.03762v5, 2017.

\bibitem{pyramid}
Xiaoyong Shen, Ying-Cong Chen, Xin Tao, Jiaya Jia. \emph{Convolutional Neural Pyramid for Image Processing.} 
arXiv:1704.02071v1 [cs.CV], 2017.

\bibitem{pressure}
Peng Su, Xiao-Rong Ding, Yuan-Ting Zhang. \emph{Long-term Blood Pressure Prediction with Deep Recurrent Neural Networks}. IEEE EMBS International Conference on Biomedical \& Health Informatics (BHI), 2018.

\bibitem{hatib}
Hatib F, Jian Z, Buddi S, Lee C, Settels J, Sibert K, Rinehart J, Cannesson M. \emph{Machine-learning Algorithm to Predict Hypotension Based on High-fidelity Arterial Pressure Waveform Analysis}. Anesthesiology. Oct;129(4):663-674, 2018.

\end{thebibliography}

% If you don't use BibTeX (why not?) , comment out or remove the previous
% line, and uncomment the following lines up to the ``}\end{bibliography}''
% line below:
%\begin{thebibliography}{99}{ %\small
% \bibitem{tag} (General form) J. K. Author, ``Name of paper,''
%   \emph{Abbrev. Title of
%   Periodical}, vol. x, no. x, pp. xxx--xxx, Abbrev. Month, year. 

% \bibitem{ito}  M. Ito et al., ``Application of amorphous oxide TFT to
%   electrophoretic display,'' \emph{J. Non-Cryst. Solids}, vol. 354, no. 19,
%   pp. 2777--2782, Feb. 2008.
  
% \bibitem{fardel}  R. Fardel, M. Nagel, F. Nuesch, T. Lippert, and
%   A. Wokaun, ``Fabrication of organic light emitting diode pixels by
%   laser-assisted forward transfer,'' \emph{Appl. Phys. Lett.}, vol. 91,
%   no. 6, Aug. 2007, Art. no. 061103.
  
% \bibitem{buncombe} J. U. Buncombe, ``Infrared navigation Part I: Theory,''
%     \emph{IEEE Trans. Aerosp. Electron. Syst.}, vol. AES-4, no. 3,
%     pp. 352--377, Sep. 1944.
      
% Uncomment the following line if you are not using BibTeX.
%}\end{thebibliography}

% LaTeX inserts the ``Address for correspondence'' heading.
\begin{correspondence}
Elise Jortberg\\
22 Cherry Hill Drive, Danvers MA, USA 01923\\
ejortberg@abiomed.com
\end{correspondence}

\end{document}